\documentclass[12pt,a4paper,twoside]{article}
\usepackage{amsmath,amssymb}
\usepackage[english]{babel}
\usepackage{caption}
\usepackage{fancyhdr}
\usepackage[T1]{fontenc}
\usepackage[pdftex]{graphicx}
\usepackage{hyperref}
\usepackage{siunitx}

\newcommand{\EatDot}[1]{}
\newcommand{\TheTitle}{Moving to VideoKifu: the last steps toward a fully automatic record-keeping of a Go game.}

\fancypagestyle{plain}{\fancyhead{}}
\hypersetup{
  pdftitle={\TheTitle},
  pdfauthor={Andrea Carta and Mario Corsolini},
  pdfsubject={\TheTitle},
  colorlinks=true,
  citecolor=[RGB]{0,80,0},
  linkcolor=[RGB]{120,0,0},
  urlcolor=[RGB]{0,0,160}}

\begin{document}

\phantomsection
\addcontentsline{toc}{section}{Front matter}
\author{Andrea Carta\thanks{~Both authors contributed equally.} \\ \url{http://www.micini.net/} \and Mario Corsolini\textsuperscript{$*$} \\ \textit{Mathematics Dept., I.T.S. ``Tullio Buzzi'' (Italy)} \\ \url{http://www.oipaz.net/}}
\title{\TheTitle}
\date{}
\maketitle

\phantomsection
\addcontentsline{toc}{section}{Abstract}
\begin{abstract}
  In a previous paper~\cite{CC15} we described the techniques we successfully employed for automatically reconstructing the whole move sequence of a Go game by means of a set of pictures. Now we describe how it is possible to reconstruct the move sequence by means of a video stream (which may be provided by an unattended webcam), possibly in real-time. Although the basic algorithms remain the same, we will discuss the new problems that arise when dealing with videos, with special care for the ones that could block a real-time analysis and require an improvement of our previous techniques or even a completely brand new approach. Eventually we present a number of preliminary but positive experimental results supporting the effectiveness of the software we are developing, built on the ideas here outlined.
\end{abstract}

\section{Introduction}

During the 2$^\text{nd}$ International Go Game Science Conference, held within the LIX European Go Congress (Liberec, CZ) on the 29$^\text{th}$--30$^\text{th}$ July 2015, we presented PhotoKifu,\footnote{~\url{http://www.oipaz.net/PhotoKifu.html}} a software for Windows$^\circledR$ operating systems indeed able to reconstruct the record of a Go game from a series of photographs taken during the game itself, each one after each move played. We described the program in detail, explaining the importance of taking each photograph immediately after a stone had been released on the goban; we described the algorithms employed to first identify the grid on the goban, also inferring the position and orientation of the camera, then to track the grid through the pictures (in order to compensate for small, accidental movements like some bumping on the table and so on), and eventually to detect every new stone placed on the goban. We also described how to avoid false negatives (stones not detected), false positives (stones wrongly detected), how to circumvent problems caused by ``disturbance'' (for example, hands of the players still visible in the pictures) as well as missing or duplicate pictures and how, in the worst cases, manual correction of the moves is allowed.
The performance of the program was very good, as shown in the paper we wrote for the occasion~\cite{CC15}, and further improved in the following releases of PhotoKifu, when we were eventually able to make use of the OpenCV\footnote{~\url{http://opencv.org/}} library, in place of the effective, but slow, ImageMagick\footnote{~\url{https://www.imagemagick.org/}} suite that was needed for some pre-processing of the pictures before the actual algorithms could do their job.

But the real reason behind the use of the OpenCV library is another one: we were already aware that a program like PhotoKifu could have just a limited success, because someone manually taking the photographs was always needed, as we could not trust the players themselves, especially during serious games, to perform such a task with the necessary care. This is not so different from having an assistant of some sort transcribing the moves, as it already happens --- much more quietly --- during the most important games.

That's why we decided, well before the Liberec conference, to switch to a video feed analysis, with the ultimate goal of writing a program able to reconstruct the record in real-time, of course without human intervention (except when an error, hopefully very uncommon, should occur). Even though we cannot expect the new program to be able to process in real-time all the frames in a video, a high percentage of them should be processed anyway, in order not to miss any move even during yose or byo-yomi periods: that's why speed becomes a major requisite, a one that would not be fulfilled without making use of the OpenCV library.\footnote{~Under the same conditions, OpenCV could be ten times or more faster than ImageMagick to pre-process a picture.}

If such a program will prove reliable it would also be possible, should a powerful enough hardware be available, to analyse several games at the same time and even broadcast the moves on the main Go servers all over the world. Kifus would be immediately available: the program itself could send or print them to interested bystanders, along with SGF files~\cite{KMH97} of every game.

Although the new program, called VideoKifu, is still under development, we can already exhibit some substantial results, as well as a lot of improvements over the last PhotoKifu version, and of course some new features specifically aimed to the analysis of a video feed (both live and deferred).

In section~\ref{sec:TrackingGrid} we describe how to locate the grid lines of a goban in the first frames of the video stream; thereafter we discuss how to swiftly follow, in the sequence of the frames, the small displacements of the grid caused by movements of the goban and/or of the camera. In section~\ref{sec:DetectingStones} we detail the many criteria used to detect the flow of the stones, closely reviewing what may go wrong. In section~\ref{sec:Conclusion} we present preliminary experimental results and we outline plausible future developments.

\section{Tracking the grid}
\label{sec:TrackingGrid}

Before starting any analysis of the stones played on the goban, it is fundamental to know where exactly the goban itself is located inside the frames extracted from the video source. That task is tackled in a way similar to the one we applied in PhotoKifu: at first, when no prior information is available, a robust yet fairly slow algorithm is used to estimate and confirm the starting location of the grid, then a fast second algorithm keeps track of accidental small variation of its location during the rest of the video.

\subsection{Starting location of the grid}
\label{subsec:StartingGrid}

The algorithm that pinpoints the initial position of the grid lines of the goban, described in~\cite[\S{2.1}]{CC15}, was fast enough for its meant purposes but it easily failed to work when more than a few dozens of stones were played. Furthermore, it was not so fast to be used in a real-time video stream. The linear Hough transform~\cite{DH72} provided by OpenCV is faster than our in-house implementation, so we switched to it, addressing the latter issue. This choice meant that we had to rewrite the entire procedure from scratch as OpenCV does not export the number of votes of the lines it recognises and our previous algorithms were deeply based on those numbers. In designing the brand new procedure described below, we took the opportunity to successfully address the former issue as well.
\begin{enumerate}
  \item\label{H2:DoG} For each frame retrieved from the video source an automatic gamma correction (based on the averages of RGB values) and a ``difference of Gaussians\footnote{~\url{https://www.imagemagick.org/Usage/convolve/\#dog}}'' filter are applied; besides, if needed, the image is scaled to a maximum of about $2$ Mpixels. Thus we obtain a B/W image --- one bit per pixel --- that highlights all the visible edges of the original frame (an example may be seen in figure~\ref{fig:Diagonal}).
  \item The linear Hough transform of the B/W image is computed and filtered: only the strongest recognised lines are kept (OpenCV sorts the lines by decreasing number of votes, even though their actual values are not returned).
  \item If there is a goban in the frame, its grid lines form two pencils\footnote{~Sets of lines through a point, which is the point at infinity (or vanishing point) of the parallel lines drawn on the surface of the goban, as shown in figure~\ref{fig:GL}.} which are transformed into points of the Hough space that are almost aligned (they are actually lying on an inverse sinusoid). Now a second run of the Hough transform is carried out in the first Hough space to identify at least a part of each one of the two pencils; as the lines selected in the previous step are about a hundred, that second transform is fast and accurate.
  \item A direct calculation shows that the coordinates in the frame of the points at infinity of each pencil are:
    \[P_\infty = \left(\rho\cos\vartheta-\frac{\sin\vartheta}{m},\rho\sin\vartheta+\frac{\cos\vartheta}{m}\right),\]
    where $(\rho,\vartheta)$ are the coordinates (in the first Hough space) of one of the lines in the pencil and $m$ is the slope of the tangent to the sinusoid at $(\rho,\vartheta)$. Knowing the points at infinity of the pencils, it is possible to evaluate the slope of the horizon and a virtual ground line $GL$ (e.g.: passing through the centre of the frame).
  \item The pencil whose lines are ``more perpendicular'' to the ground line is selected and labelled as the ``vertical'' pencil (the other being of course the ``horizontal'' one). All the intersections among $GL$ and the ``vertical'' lines are computed, in order to evaluate their median distance, which is constant for the lines of the grid, as shown in figure~\ref{fig:GL}.
    \begin{figure}[!htb]
      \centering
      \setlength{\unitlength}{0.625\linewidth}
      \begin{picture}(1.600,0.687)
        \put(0,0){\includegraphics[width=1.600\unitlength]{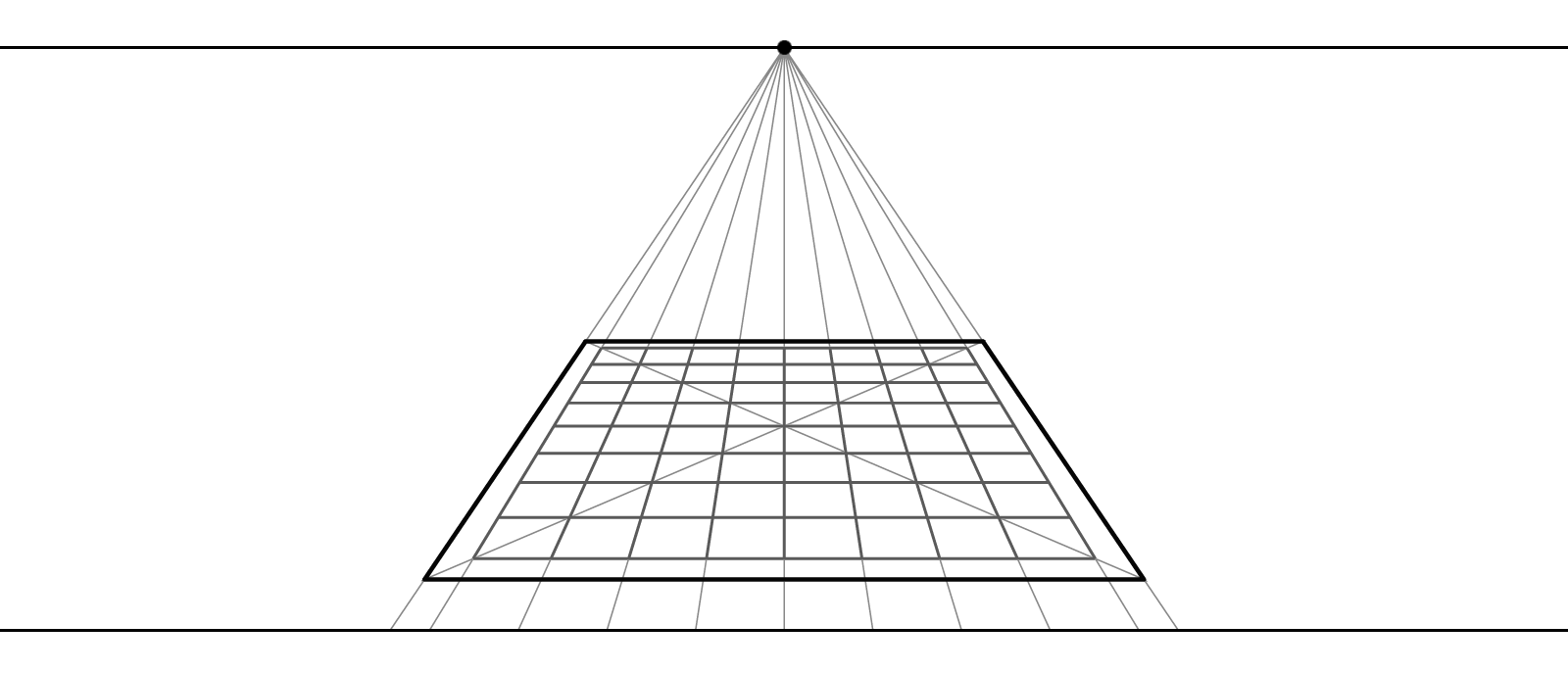}}
        \put(0.800,.673){\makebox(0,0){$V$}}
        \put(0.010,.593){Horizon}
        \put(0.397,.015){\makebox(0,0){$A$}}
        \put(0.437,.015){\makebox(0,0){$1$}}
        \put(0.528,.015){\makebox(0,0){$2$}}
        \put(0.619,.015){\makebox(0,0){$3$}}
        \put(0.709,.015){\makebox(0,0){$4$}}
        \put(0.800,.015){\makebox(0,0){$5$}}
        \put(0.890,.015){\makebox(0,0){$6$}}
        \put(0.981,.015){\makebox(0,0){$7$}}
        \put(1.072,.015){\makebox(0,0){$8$}}
        \put(1.163,.015){\makebox(0,0){$9$}}
        \put(1.203,.015){\makebox(0,0){$B$}}
        \put(0.010,.056){$GL$}
      \end{picture}
      \caption{$V$ is the point at infinity of the ``vertical'' lines; as the ``horizontal'' ones are perfectly horizontal, their point at infinity is truly at infinity and the horizon is a horizontal line (although this is a simplified particular case, it does not substantially differ from the general one). On the ground line $GL$, which may be any line parallel (but not coincident) to the horizon, the points from $1$ to $9$ are each other equidistant, a peculiarity which does not belong to the intersections between the ``vertical'' lines and any other line not parallel to the horizon, as the two diagonals.}
      \label{fig:GL}
    \end{figure}
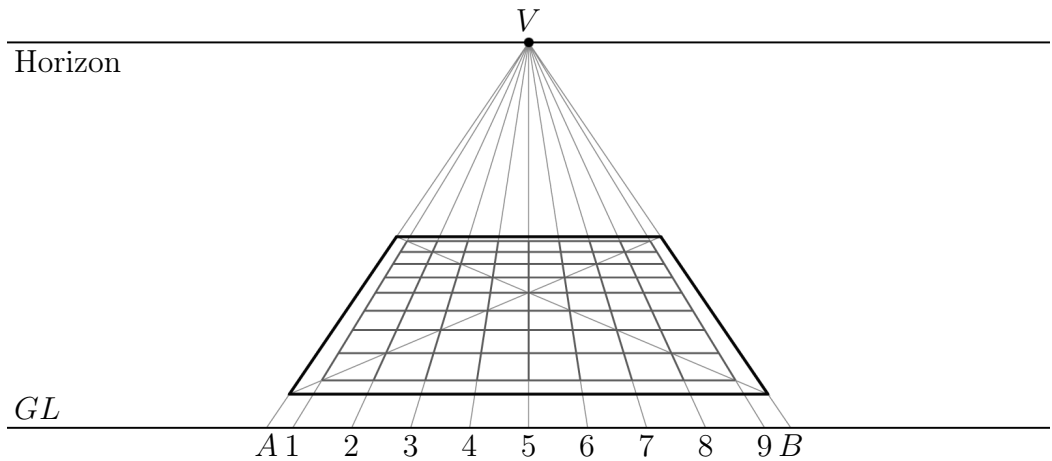
  \item\label{H2:filtering} If some lines are missing from the ``vertical'' pencil (because they are concealed by stones or other disturbances) they may be added at this point. On the contrary, lines whose intersections with $GL$ are significantly different from an integer multiple of the median distance are pruned.
  \item The best equispaced subset of $19$ lines\footnote{~Or $13$, or $9$, according to which is the biggest available subset.} is selected as the one forming the ``vertical'' lines of the grid of the goban.
  \item ``Vertical'' lines are sorted from left to right, the ``horizontal'' ones from top to bottom, then each ``vertical'' line is tidily coupled with a ``horizontal'' one, in consecutive order. After that another linear Hough transform is computed among the intersections of each couple, for the sake of selecting the longest aligned subset: those intersections should constitute a diagonal of the grid, as shown in figure~\ref{fig:Diagonal}. Since the intersections are at most $19$, the transform is computed extremely fast.
    \begin{figure}[!htb]
      \setlength{\unitlength}{.625\linewidth}
      \begin{picture}(1.600,1.080)
        \put(0,0){\frame{\includegraphics[width=1.600\unitlength]{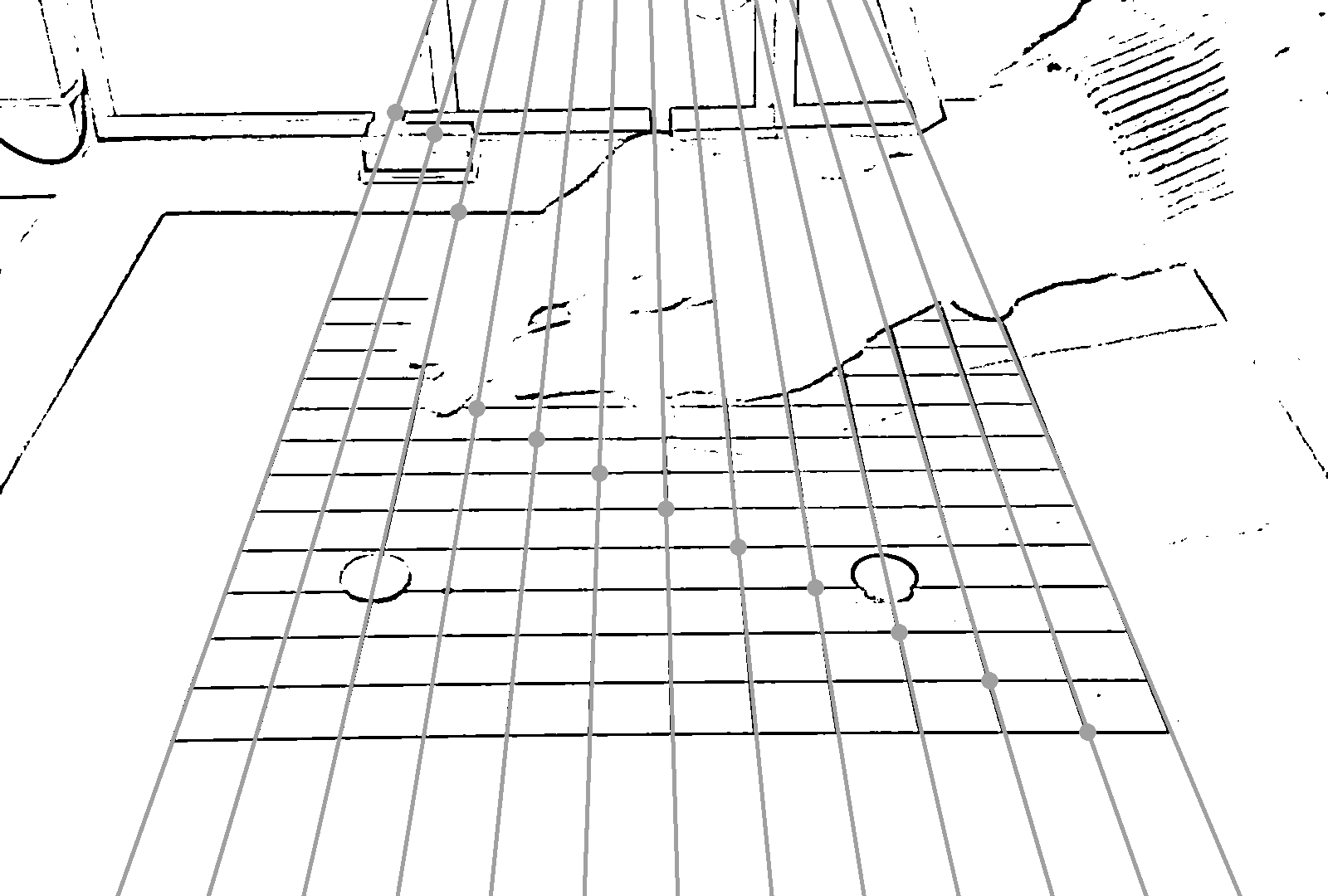}}}
        \multiput(0.486,.954)(0,.0005){5}{\color{white}$1$}
        \multiput(0.533,.928)(0,.0005){5}{\color{white}$2$}
        \multiput(0.562,.834)(0,.0005){5}{\color{white}$3$}
        \multiput(0.584,.597)(0,.0005){5}{\color{white}$4$}
        \multiput(0.656,.560)(0,.0005){5}{\color{white}$5$}
        \multiput(0.732,.519)(0,.0005){5}{\color{white}$6$}
        \multiput(0.812,.476)(0,.0005){5}{\color{white}$7$}
        \multiput(0.899,.430)(0,.0005){5}{\color{white}$8$}
        \multiput(0.992,.381)(0,.0005){5}{\color{white}$9$}
        \multiput(1.093,.327)(0,.0005){5}{\color{white}$10$}
        \multiput(1.202,.269)(0,.0005){5}{\color{white}$11$}
        \multiput(1.320,.207)(0,.0005){5}{\color{white}$12$}
        \multiput(0.485,.955)(.0005,0){5}{\color{white}$1$}
        \multiput(0.532,.929)(.0005,0){5}{\color{white}$2$}
        \multiput(0.561,.835)(.0005,0){5}{\color{white}$3$}
        \multiput(0.583,.598)(.0005,0){5}{\color{white}$4$}
        \multiput(0.655,.561)(.0005,0){5}{\color{white}$5$}
        \multiput(0.731,.520)(.0005,0){5}{\color{white}$6$}
        \multiput(0.811,.477)(.0005,0){5}{\color{white}$7$}
        \multiput(0.898,.431)(.0005,0){5}{\color{white}$8$}
        \multiput(0.991,.382)(.0005,0){5}{\color{white}$9$}
        \multiput(1.092,.328)(.0005,0){5}{\color{white}$10$}
        \multiput(1.201,.270)(.0005,0){5}{\color{white}$11$}
        \multiput(1.319,.208)(.0005,0){5}{\color{white}$12$}
        \put(0.486,.955){$1$}
        \put(0.533,.929){$2$}
        \put(0.533,.929){$2$}
        \put(0.562,.835){$3$}
        \put(0.584,.598){$4$}
        \put(0.656,.561){$5$}
        \put(0.732,.520){$6$}
        \put(0.812,.477){$7$}
        \put(0.899,.431){$8$}
        \put(0.992,.382){$9$}
        \put(1.093,.328){$10$}
        \put(1.202,.270){$11$}
        \put(1.320,.208){$12$}
      \end{picture}
      \caption{the twelve numbered dots represent the intersection between the ``vertical'' grey lines selected by the algorithm and the longest ``horizontal'' lines found in the Hough space (not highlighted in the figure). It is clear that the longest aligned subset of grey dots is from $4$ to $12$: those points indeed form part of an actual diagonal of the grid.}
      \label{fig:Diagonal}
    \end{figure}
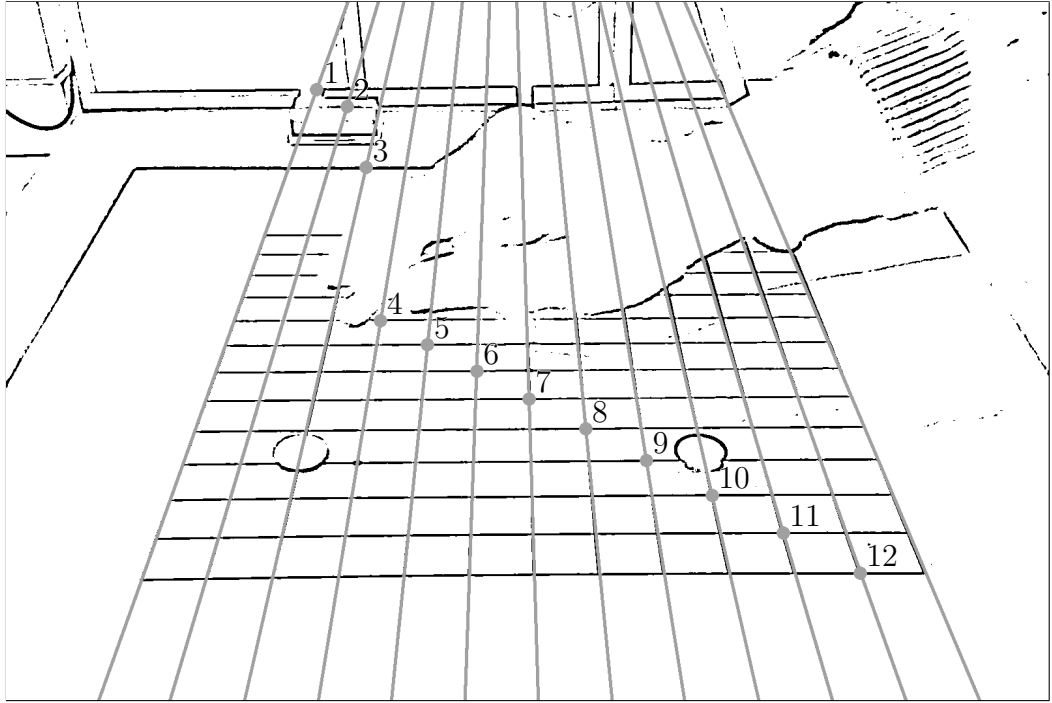
  \item Knowing from the previous step a subset of consecutive ``horizontal'' lines, it is now possible to act as in step~\ref{H2:filtering} to obtain the complete set of ``horizontal'' lines of the grid.
  \item Sometimes, depending on the angle of view and the thickness of the goban, the lower wooden border of the board appears in the frame exactly at the position where a grid line could have been. Therefore it is necessary to check if such a border has been mistaken for a grid line and to amend it, if that was the case. This is accomplished by evaluating the actual presence in the original frame of real intersections along the borders of the computed grid.
\end{enumerate}

Unless differently stated, every timing mentioned throughout this paper will be referred to a personal computer equipped with an Intel$^\circledR$ Core{\texttrademark} i7-4770 CPU @ 3.40GHz and integrated HD Graphics 4600. On that machine the entire procedure usually takes less than $0.3$ seconds\footnote{~Strong background noise may considerably increase that amount of time.} to be applied to a 1080p frame, thus enabling its use in real-time during a video feed. Moreover the new algorithm is able to locate the grid even if it is partially concealed: experiments showed that it works most of the times when at least $50\%$ of the grid area is visible (it sometimes works even in worse conditions, but the percentage of success quickly drops). This allows the use of the algorithm up to the mid-endgame, to rectify errors that may sporadically occur in the automatic micro-recalibration procedure, as discussed next.

\subsection{Automatic micro-recalibration}
\label{subsec:AMR}

In~\cite[\S{2.2}]{CC15} we described a triple approach to the problem of tracking small movements of the grid between consecutive pictures. Of course the same methods could have been used even on a video stream, and we actually maintained some of the fundamental ideas of the old approach; yet we chose to discard the use of any kind of Hough transform (either corner, linear or elliptic) in order to achieve higher speed of execution.\footnote{~The new procedure is usually executed on a 1080p frame in less than 0.02 seconds.} Thanks to the use of the MatchTemplate function supplied by OpenCV (which comes with a useful standardised index of reliability), we could merge two of the previous methods, actually improving both of them.

First of all we apply to each frame of the video the same filters as described in step~\hyperref[H2:DoG]{\ref*{subsec:StartingGrid}.\ref*{H2:DoG}}. When the location of the grid is confirmed, we cut four small portion of the filtered image around the four corners of the grid and we save them.

\begin{figure}[!htb]
  \includegraphics[width=\linewidth]{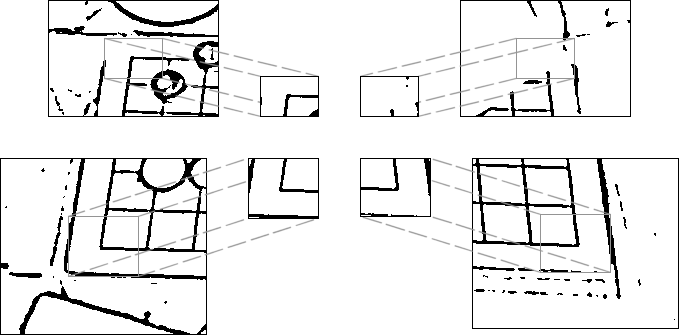}
  \caption{the small images on the centre are the templates cut around each of the four corners in the previous frame; the large images are the areas where the templates are searched for in the current frame. In the upper right corner a player's arm partly conceals the corner, yet the few existing features of the template are recognised by MatchTemplate and the corner is correctly located, even if with a low reliability of $18.2\%$ (and so our algorithm will discard it), whereas the other corners, obviously perfectly pinpointed, have indexes of $96.4\%$, $92.3\%$ and $87.0\%$ (counterclockwise).}
  \label{fig:AMR}
\end{figure}

After the following frame is loaded, MatchTemplate compares the previously saved templates against the new filtered frame, looking for matches (as shown in figure~\ref{fig:AMR}) that, if found, define the new corners of the grid in the new frame, allowing the process to be iterated in the subsequent frames. That allows to track at the same time both visible corners and the ones hidden by stones, the features of which are sought by MatchTemplate.

A failure may occur when a corner is temporary hidden by a disturbance, or when a stone is played between two consecutive frames in a previously empty corner but, as stones are played one at a time, most of the times three corners should be recognisable, which are enough to geometrically reconstruct the entire grid.

Experiments show that this new method is fast and reliable, but it may suffer from a drift phenomenon. E.g.: if a player's hand slowly moves above one of the corners, that corner may be ``dragged'' around. We fixed that evaluating a linear regression of the black pixels\footnote{~Those computations are applied to frames filtered as the one shown in figure~\ref{fig:Diagonal}.} in the neighbourhood of each line joining two consecutive estimated corners: this calculation replaces the powerful but slower linear Hough transform used in PhotoKifu. The area in which the actual line is searched for is very thin, so accuracy of the results is comparable or even higher than formerly, as for time reasons the Hough transform had to be evaluated with a low angle resolution.

\section{Detecting the stones}
\label{sec:DetectingStones}

The idea behind VideoKifu is simple: every single frame of the video feed, live or deferred, may be analysed by means of the same algorithm we employed in PhotoKifu, effectively turning the video in a series of photographs. Of course, most frames would show no changes at all, a situation that in PhotoKifu would have prompted an error, asking the user if the photograph really was a duplicate of the previous one; in VideoKifu such frames could simply be skipped. In fact, every now and then a new stone would appear on the goban, and in this case only, should the algorithm detect it, there would be an update of the current position. Those updates could go on until the end of the game, if the algorithm proves fast enough to successfully process most frames: if this is not the case, some moves could be missed should the pace of the game go frenzy (for example during byo-yomi).
But these updates, too, are not a simple matter. On the goban, stones do not appear from nowhere: first they are carried to the grid intersection where they are intended to be released, then they are placed there, and eventually the player's hand retreats from the goban. It means that on each move there are about 2 seconds during which the goban will be heavily ``disturbed'', no frame could be identical to the previous one and the stone could nonetheless, or could not, be recognizable.
Two seconds of ``disturbance'' could easily make a false positive appear, and in such situations a human intervention will be needed soon, before the analysis of the video could go awry and eventually stop completely.
Although the algorithm employed in PhotoKifu performs very well in case of ``disturbance'', false positives cannot be ruled out; also, should the stone be clearly visible when still in the player's hand, it could be detected before being released on the goban. Captured stones, for a similar reason, are another big issue: if the program removes them immediately after the capturing stone has been detected, false positives could --- and would --- occur, because the removed stones' colour is the one expected to move; if the program waits instead for the stones to be really removed, in order not to misunderstand them for false positives, then PhotoKifu's algorithm cannot be of great use, because it has been designed to detect stones, not empty intersections.

\subsection{Stability}

As previously noted, any video feed of a Go game will show a lot of ``disturbed'' frames, because the hands (and possibly the whole arms) of the players will be visible immediately before and after the placement of a stone, for a time that can be approximately estimated in at least two seconds for each move. How many frames are we talking about? In a typical 19$\times$19 Go game about 250 moves are played, accounting for 500 seconds of ``disturbance'': if the program is able to analyse at least 5 or 6 frames per second,\footnote{~Indeed we reached 6 frames per second for 1080p video streams.} it will have to deal with 2500 ``disturbed'' ones in the whole game.
As reported in~\cite{CC15}, PhotoKifu is able, under good conditions, to detect 100\% of the stones even in the presence of disturbance, given they still can be seen; but as PhotoKifu only analyses one photograph per move it's unlikely that the number of pictures affected by disturbance will exceed one hundred or so.\footnote{~During our tests we counted 63 at worst.} That's almost nothing when compared to the several thousands which, according to the previous estimate, VideoKifu will deal with, and it's quite obvious that despite the algorithm looking flawless it is likely that a false positive will occur, sooner or later. Again, we must remember that a single one could prove fatal for the analysis of a live video feed.

That's why we introduced the concept of stability: it means that no position could be validated before it occurs a given number of consecutive times. It is difficult to determine how many times, because although it's quite clear that ``the most, the best'', the more they are, the more likely it will be to miss a move when the pace is fast, because in such a case there could be not enough consecutive frames devoid of disturbance.
What is probably a good compromise is a ``stability time'' no longer than one second and no shorter than half a second; it means that, for example, if the program is able to analyse 6 frame per second, it should wait three to six frames before a position --- if it doesn't change in between --- can be validated and the record updated.
So, even if a false positive occurs, as most disturbance is caused by the hands of the players moving on the goban, it's quite obvious that the movement itself will make it impossible for the error to occur again: and even if it would, it won't be the same one, but most likely will concern another intersection.
Of course, if the frames won't show any disturbance, stability won't matter: the appearance of the goban, once the last stone played has been detected, won't change, and no update will ever be necessary.

The only drawback of the stability concept is that, despite choosing carefully the number of frames to wait for, if the pace is really fast it's still possible to miss a stone: indeed, when the pace is fast a player may put a stone on the goban when the opponent's hand is still retreating, preventing the stability to occur. By the way, this is not a big issue, as we were already able in PhotoKifu to detect two moves at once, also establishing the right order; three and even four moves could be similarly detected, although in such cases the right order could only be guessed, something that sooner or later the user should check and possibly correct.
Such occurrences, by the way, almost never happen: all the tests we have performed by now show that even at a fast pace (e.g.: ten minutes for the whole game on a 13$\times$13 goban) it's high unusual for two moves to be detected together; detecting three of four stone at the same time is something that has never occurred (unless a stone is not detected immediately, but that's another matter). So it's safe, at the moment, to state that enforcing the stability concept is all we need to solve most problems caused by disturbance.

\subsection{Tangential Hough transform}

Unfortunately there are many kind of disturbance. Although in most cases it is caused by the players' hands carrying the stones on the goban, it's not uncommon to see the players leaning over, gesticulating, holding out portable phones or tablets and so on. If in such cases they remain still for some time, and a false positive occurs, the frames following the error will look almost identical, possibly triggering the stability's requirements and eventually validating a wrong position.
In order to prevent such an occurrence we strengthened PhotoKifu's algorithm forcing it to compute the elliptical Hough transform for every stone possibly detected.\footnote{~In PhotoKifu the stones are detected by means of a complex function that relies on known stones' data, and only when the function's output is uncertain the Hough transform is also computed.} This simple trick prevented most false positives, that indeed would have occurred when the players leaned over the goban (especially if wearing black or white clothes): but on rare occasions even the Hough transform proved useless. For example, during the 2015 Pisa International Go Tournament we filmed the game between Andrea Pignelli and Antonio Albano (both ranked 6~kyu EGF then); after 34 moves Pignelli leaned over the goban's upper left corner, staying perfectly still for about two minutes: despite the continuous use of the Hough transform a false positive was soon detected and validated (approximately after 50 frames) and we had to improve the transform itself in order to resolve the problem.

How exactly we implemented the transform in PhotoKifu is described with the help of the following figures, which show some real cases.

\begin{figure}[!htb]
  \frame{\includegraphics[width=0.3\linewidth]{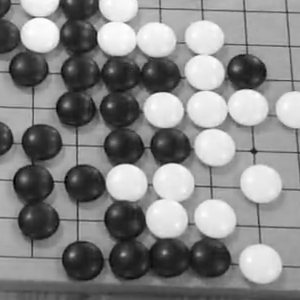}}
  \hfill
  \frame{\includegraphics[width=0.3\linewidth]{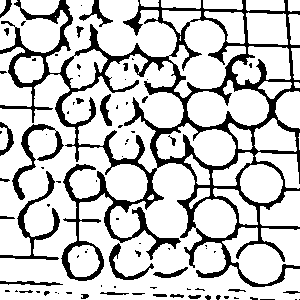}}
  \hfill
  \frame{\includegraphics[width=0.3\linewidth]{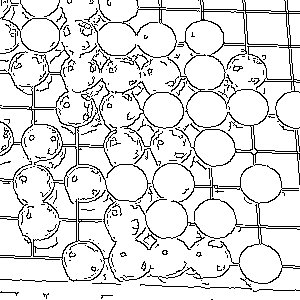}}
  \caption{a fragment of the original frame on the left, ``difference of Gaussians'' filtering on the centre and Canny filtering on the right.}
  \label{fig:Pignelli-Albano}
\end{figure}

First of all we apply the Canny filter~\cite{Can86}, with Otsu threshold~\cite{Ots79}, to the currently analysed frame. Figure~\ref{fig:Pignelli-Albano} shows why we chose to apply an edge detector different from the ``difference of Gaussians'' already evaluated during the automatic micro-recalibration (as explained in section~\ref{subsec:AMR}).

Both the filters exhibit noise inside the black stones, whose borders are more irregular than the white's ones, but those are not hindrances. The real problem with ``difference of Gaussians'' is that black stones appear thicker and noticeably smaller than the white ones, so Canny is more advisable for the current purpose. On the other hand when using Canny the grid lines are transformed into a couple of thin contiguous lines, whilst ``difference of Gaussians'', if used with appropriate radius and sigmas, preserves them as single bold lines, which is preferable for the sake of locating the grid.

Once the filtered image is obtained, the stones' diameter is estimated, on the grounds of the distances between the grid lines. Because the stone we are looking for could be out of centre, we select a square, its side about half\footnote{~In theory the side could be up to an entire stone's diameter. Future experiments will set the most effective value.} a stone's diameter, around the intersection of interest; inside such square we search for the stone's central point,  Finally, for each point inside the square, a circular corona, whose mean diameter approximates that of a stone, is scanned for dark pixels.

\begin{figure}[!htb]
  \begin{minipage}{0.3\linewidth}
    \frame{\includegraphics[width=\linewidth]{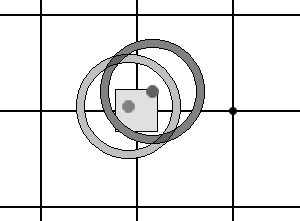}}
    \caption{}
    \label{fig:Coronas}
  \end{minipage}
  \hfill
  \begin{minipage}{0.3\linewidth}
    \frame{\includegraphics[width=\linewidth]{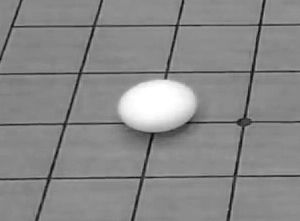}}
    \caption{}
    \label{fig:Real}
  \end{minipage}
  \hfill
  \begin{minipage}{0.3\linewidth}
    \frame{\includegraphics[width=\linewidth]{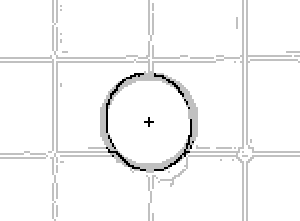}}
    \caption{}
    \label{fig:RealHough}
  \end{minipage}
\end{figure}

Figure~\ref{fig:Coronas} shows how the research process works: the square marks the area surrounding a specific intersection, where a stone is supposed to be and its centre is looked for, and for two specific points the related coronas are shown; a light one, for a point close to the intersection, and a dark one, for a point close to the border of the square.\footnote{~Perspective effects are neglected in this figure.} When all the points inside the square have been checked, the one whose corona contains the maximum number of dark pixels is selected, given the number of these pixels is greater than a fixed threshold.

While a classical circular Hough transform only searches for pixels lying on a circumference,\footnote{~We discarded the use of an elliptical Hough transform as it was too time consuming.} this variation looks for them inside a circular corona: the reason can be easily inferred by looking at a real case, as depicted in figures~\ref{fig:Real} and~\ref{fig:RealHough}. Figure~\ref{fig:Real} shows a lonely stone lying on a goban, and figure~\ref{fig:RealHough} shows what happens after the perspective is straightened and a Canny filter applied. The small cross shows the centre of the stone as detected by the transform,\footnote{~A vertical shift, due to perspective effects, is visible. During the analysis we use PhotoKifu's same algorithm to approximate the location and the orientation of the observer's point of view in order to compensate for that.} that is the centre of the corona which contains the maximum number of dark pixels (the corona is also shown):\footnote{~From now on we won't mention any more that the pixels we are looking for must be the dark ones.} as it's now obvious, despite the perspective's straightening the stone's borders do not take on a circular shape, due to shadows, small reflections and above all the fact that stones are not flat; they have a lenticular shape that may alter their projected image especially if the point of view is low. That's why looking for pixels lying on a circumference would miss most of them, and only expanding the searching area it is possible to collect them all, as clearly visible in figure~\ref{fig:RealHough} despite the stone looking elliptical.
\begin{figure}[!htb]
  \begin{minipage}{0.3\linewidth}
    \frame{\includegraphics[width=\linewidth]{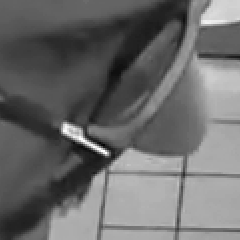}}
    \caption{}
    \label{fig:Nose}
  \end{minipage}
  \hfill
  \begin{minipage}{0.3\linewidth}
    \frame{\includegraphics[width=\linewidth]{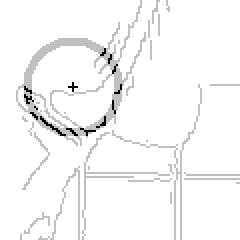}}
    \caption{}
    \label{fig:NoseHough}
  \end{minipage}
  \hfill
  \begin{minipage}{0.3\linewidth}
    \frame{\includegraphics[width=\linewidth]{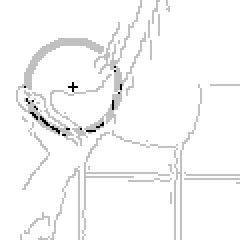}}
    \caption{}
    \label{fig:NoseHoughNew}
  \end{minipage}
\end{figure}

The efficiency of this transform is high, but there is a drawback: in the presence of disturbance many pixels which do not belong to a stone's border could turn up inside the corona. That's what happened during the game we described earlier, when a player remained still on the goban for about 2 minutes and a false positive was detected despite the transform. Figures~\ref{fig:Nose} and~\ref{fig:NoseHough} show the reason: in figure~\ref{fig:Nose} we see the player's nose and glasses hiding the goban's corner, in figure~\ref{fig:NoseHough} is shown the situation after the filtering. A lot of pixels, all belonging to the glasses, turn up inside the ``best'' corona --- the one that contains most pixels --- and they are so many that the threshold is passed; increasing its value is not useful as it prevents stones whose borders are not so neat, as it usually happens when the goban is crowded, to be detected.

In order to solve the problem a comparison between figure~\ref{fig:RealHough} and figure~\ref{fig:NoseHough} is needed: it is quite obvious that in the first case --- a real stone --- all the pixels found inside the corona really are part of a circumference, while in the second one --- disturbance --- they are scattered almost randomly. Therefore it is mandatory to differentiate between the two situations, and that's why we decided to check, for each pixel, a property we call ``tangentiality'': a pixel is defined ``tangential'' if it's part of an arc whose centre is close to the corona's central point.
If this is not true, we can assume the pixel is part of some disturbance instead and can be neglected. Figure~\ref{fig:NoseHoughNew} shows what happens when the ``tangentiality'' of figure~\ref{fig:NoseHough} pixels is checked: tangential ones are dark, while all the others (that we define ``radial'') are light.

\begin{figure}[!htb]
  \begin{minipage}{.35\linewidth}
    \frame{\includegraphics[width=\linewidth]{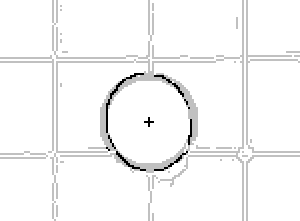}}
    \caption{}
    \label{fig:RealHoughNew}
  \end{minipage}
  \hfill
  \begin{minipage}{.6\linewidth}
    \setlength{\unitlength}{1.388889\linewidth}
    \begin{picture}(.720,.309)
      \put(0,0){\frame{\includegraphics[width=.720\unitlength]{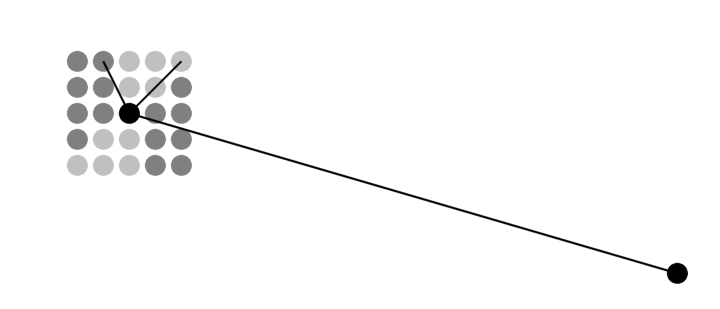}}}
      \put(0.098,.147){\makebox(0,0){\rotatebox{60}{$\xrightarrow{\makebox[2em]{}}$}}}
      \put(0.066,.088){\makebox(0,0){$P$}}
      \put(0.181,.281){\makebox(0,0){$P_1$}}
      \put(0.103,.281){\makebox(0,0){$P_2$}}
      \put(0.687,.071){\makebox(0,0){$C$}}
    \end{picture}
    \caption{}
    \label{fig:Scheme}
  \end{minipage}
\end{figure}

As we can see, only half of all pixels inside the corona are tangential: without the radials' contribution their number cannot now pass the threshold and that's what eventually prevents a false positive to appear. For comparison, almost all the pixels in figure~\ref{fig:RealHough} are instead tangential, as shown in figure~\ref{fig:RealHoughNew}: that's because a stone, not some kind of disturbance, is indeed there. Therefore the number of pixels needed to pass the threshold is reached and the stone detected.

How to determine if a pixel is tangential is shown in figure~\ref{fig:Scheme}.

If point $C$ marks the assumed centre of the stone, and we are checking the tangentiality of pixel $P$, supposedly found inside the searching corona, we check all pixels $P_n$ in a 5$\times$5 square around $P$ (only if such pixels are dark and are themselves inside the corona), computing for each one the value of the angle between vectors $P-C$ and $P-P_n$. If the angle is between $\ang{60}$ and $\ang{120}$ we mark $P_n$ as ``good'' (light pixels in figure~\ref{fig:Scheme}), otherwise as ``bad'' (dark pixels in figure~\ref{fig:Scheme}). If the number of good points exceeds the number of bad points we define pixel $P$ ``tangential'', otherwise ``radial''.

We could not test every single frame among the ones we have available, as their number exceeds hundreds of thousands, so we only scrutinized some selected ones, in which a lot of stones are present --- a situation that worsens the problems encountered by the Hough transform, as contiguous stones' borders tend to cancel each other out: nevertheless these limited tests confirmed that adding tangentiality to the Hough transform improves so much its efficiency that it almost matches PhotoKifu's main function's accuracy\footnote{~Compare~\cite[\S{4}]{CC15}.} (in most frames only one stone among 200 was not detected, or wrongly detected), and can be indeed relied upon in ruling out any kind of disturbance.

\subsection{The empty intersection problem}

\begin{figure}[!htb]
  \begin{minipage}{.475\linewidth}
    \frame{\includegraphics[width=\linewidth]{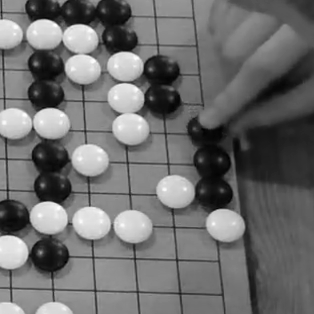}}
    \caption{}
    \label{fig:T13}
  \end{minipage}
  \hfill
  \begin{minipage}{.475\linewidth}
    \frame{\includegraphics[width=\linewidth]{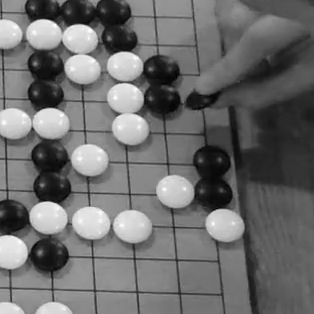}}
    \caption{}
    \label{fig:T14}
  \end{minipage}
\end{figure}

During the aforementioned game between Andrea Pignelli and Antonio Albano, the latter played the 145$^\text{th}$ move on T13, as shown in figure~\ref{fig:T13}, but hesitated a moment too much before releasing the stone: hence, as shown in figure~\ref{fig:T14}, it was wrongly detected on T14, because it was clearly visible over this intersection as long as needed.

What happened when the stone was eventually released on T13? Initially the program replied with ``detected a stone whose colour is wrong'', because after detecting a black stone on T14, it was now expecting a white one. Such a problem, despite occurring just once in all the games we recorded, should not be uncommon: for example, it would certainly occur each time a player would place a stone on the goban, then change his mind and push it, without releasing the grip, on another intersection close by.

The only way to deal with such a problem is to check, in such cases, if the stone previously played has disappeared, and act accordingly. That's even more necessary because a similar situation may occur each time one or more stones are captured and removed from the goban: if the program removes them immediately after the capturing stone's detection, without waiting for them to really disappear, it's possible that one of them will be again detected and recorded as it were the expected move (as it's now the captured stone's colour turn to move).\footnote{~That could be avoided only if there is only one stone captured, and suicide is forbidden.} Both situations require an algorithm capable of recognizing when an intersection, previously occupied by a stone, becomes empty.\footnote{~PhotoKifu is not affected by this problem, as users are advised to take pictures only after each move has been carried out, removal of any captured stones included.}

Developing such an algorithm is not easy, as the whole idea behind programs such as PhotoKifu and VideoKifu is ``detecting a stone placed on a previously empty intersection'', not the opposite; furthermore, it is imperative not only to detect the empty intersection, but also to tell it apart from any kind of disturbance: were this not the case, a hand passing over the stone expected to be removed could be misunderstood for the empty intersection, making completely useless every effort to avoid the error we described before.

Although not uncommon, neither such cases occur frequently: that's why we are still working on the algorithm that should tell apart empty intersections from stones. At the moment, these steps look promising:
\begin{enumerate}
  \item computing, for all intersections known to be empty, the average values of uniformity, chrominance, hue, difference with mean white stone / mean black stone / mean empty intersection;\footnote{~The values to be averaged are already known for every intersection, as are previously computed by PhotoKifu's main function, whose goal is detecting the stones.}
  \item computing also, for each one of these average values, the standard deviation;
  \item retrieving (from PhotoKifu's main function) these same values for the intersection to be checked, the one covered by the stone expected to disappear;
  \item verifying which ones, among these values, are in the range defined by the average values computed above $\pm$ two times the standard deviation;
  \item if every one of these values --- six in all --- lies in the requested range, we assume the intersection has become empty and act accordingly.
\end{enumerate}

Extensive tests are difficult to undertake, given the huge number of frames to examine (the aforementioned game between Pignelli and Albano has more than 268,000), but something can be done by means of the pictures taken for PhotoKifu's analyses.\footnote{~These pictures are online at PhotoKifu's homepage.} For example, in the game between Alberto Zingoni and Alessandro Pace 238 pictures were taken, and another 277 covered the game between Alessandro Pace and Pasquale De Lucia. In these 515 pictures approximately 65,000 stones are present, and the algorithm described before, when tested on them, only goes wrong 107 times (less than 0.17\%), most of them in a few pictures affected by heavy disturbance: this means that 107 times out of 65,000 a stone was misunderstood for an empty intersection, and this in turn means it becomes virtually impossible, when stability is taken into account, to erroneously remove a stone that has not disappeared.
In the course of these two games 50 stones were captured: 45 times (90\%) the algorithm correctly acknowledged the stone had disappeared, a percentage good enough to make us feel confident that, given enough frames, all stones bound for removal would indeed be removed.\footnote{~For example, if three consecutive, identical frames were needed to achieve stability (that means half a second if the video analysis can process 6~FPS), no more than six would be required in 95\% of the cases.} Furthermore, it's possible to greatly improve this percentage if we loosen a bit the criteria required to decide if an intersection is empty: indeed, if we only require five, instead of all six, values to lie in the requested range, the percentage reaches a remarkable 96\% (on the two games previously mentioned).\footnote{~If such were the case, stability will be reached in six or less frames 99\% of the times, if three consecutive ones are needed.} Of course, were this the case, the number of stones misunderstood for empty intersections also would rise, but not dramatically: 476 out of 65,000, still much less than 1\%, a percentage unlikely to induce any errors, once stability is taken into account.

Once the best criterion will be chosen --- further tests are needed --- we are extremely confident that this algorithm will indeed be capable to correctly detect and remove all these stones that really have disappeared from the intersections which they were previously placed over, whatever the reason. We are also confident that in most games the improvements depicted thus far should be enough to let a video analysis reach the end without problems, as all tests performed until now suggest.

\section{Conclusion}
\label{sec:Conclusion}

As of today we have seven games recorded; one of them, the friendly one between us, was recorded twice, from both sides of the goban, by means of a DSLR camera and a tablet. Another game was analysed live by means of a webcam and also recorded, frame by frame, by the program itself during the analysis; so it too appears twice in table~\ref{table:Videos}, in which we summarize the games' data.

\begin{table}[!htb]
  \setlength\tabcolsep{3.5pt}
  \centering
  \begin{tabular}{|l||c|c|r|c|r|}
    \hline
    \rule{1.218em}{0pt}Game                        &Moves&    Resolution    &     Duration     &  FPS  & Frames\\
                                                   &     &                  &                  & Re/Pr & Re/Pr\rule{.2em}{0pt}\\
    \hline
    \hline
    1. Carta-Corsolini:                            &  96 &  640$\times$480  &          0:16:33 & 25    &  24,844\\
    \rule{1.218em}{0pt}friendly game, 13$\times$13 &     &      DSLR        &                  &  6.25 &   6,211\\
    \hline
    2. Carta-Corsolini:                            &  96 & 1920$\times$1080 &          0:16.33 & 30    &  29,795\\
    \rule{1.218em}{0pt}friendly game, 13$\times$13 &     &      tablet      &                  &  6    &   5,959\\
    \hline
    3. Pignelli-Albano:                            & 233 & 1440$\times$1080 &          3:06:07 & 24    & 268,017\\
    \rule{1.218em}{0pt}Pisa 2015                   &     &      tablet      &                  &  6    &  67,004\\
    \hline
    4. Pantalone-Balzaretti:                       & 143 & 1920$\times$1080 & $\approx$1:10:00 &  4    &  16,740\\
    \rule{1.218em}{0pt}Pisa 2016, live             &     &      webcam      &                  &  4    &  16,740\\
    \hline
    5. Pantalone-Balzaretti:                       & 143 & 1920$\times$1080 &          0:46:30 &  6    &  16,740\\
    \rule{1.218em}{0pt}Pisa 2016, deferred         &     &      webcam      &                  &  6    &  16,740\\
    \hline
    6. De Lazzari-Greenberg:                       & 231 &  640$\times$480  &          1:07:36 & 31.6  & 128,093\\
    \rule{1.218em}{0pt}Pisa 2016                   &     &      smartphone  &                  &  6    &  24,336\\
    \hline
    7. Ragno-Gioia:                                & 190 &  640$\times$480  &          1:14:21 & 15    &  66,947\\
    \rule{1.218em}{0pt}Pisa 2016                   &     &      smartphone  &                  &  5    &  22,316\\
    \hline
    8. Telesca-Metta:                              & 260 & 1920$\times$1080 &          0:43:29 & 30    &  78,274\\
    \rule{1.218em}{0pt}Pisa 2016                   &     &      tablet      &                  &  6    &  15,654\\
    \hline
    9. Martinelli-van den                          & 262 & 1280$\times$720  &          1:51:35 & 15    & 100,469\\
    \rule{1.218em}{0pt}Busken: Roma 2016           &     &      smartphone  &                  &  5    &  33,490\\
    \hline
  \end{tabular}
  \caption{the analysed videos, available online at PhotoKifu's homepage. In the two rightmost columns the upper numbers indicate the (real) values saved in the video, the lower ones are pertinent to the actual frames processed by our program during the tests.}
  \label{table:Videos}
\end{table}

At the moment the analyses work as follows:
\begin{enumerate}
  \item Carta-Corsolini (DSLR camera): no problems at all.
  \item Carta-Corsolini (tablet): no problems at all.
  \item Pignelli-Albano:\footnote{~\url{http://www.europeangodatabase.eu/EGD/Tournament_Card.php?&key=T150306B}} no problems at all.
  \item Pantalone-Balzaretti\footnote{~\url{http://www.europeangodatabase.eu/EGD/Tournament_Card.php?&key=T161022B}\label{fn:Pisa2016}} (live): the second stone (of course a white one) was believed to be black, because of a bug in the program (now fixed); it was possible to correct the problem on the fly and proceed to the end without further inconveniences. The low value of FPS was caused by the use of a portable computer, slower than our machine of reference.
  \item Pantalone-Balzaretti\textsuperscript{\ref{fn:Pisa2016}} (deferred): no problems at all. The different duration between this video and the live one is caused by the fact that live video was saved declaring 6~FPS although they actually were only 4~FPS.
  \item De Lazzari-Greenberg:\textsuperscript{\ref{fn:Pisa2016}} there are difficulties in locating the grid, but no further problems until the end of the game. However, moves 165--166 and 167--168 are inverted, as the program cannot detect moves 165--166 first (the pace was too fast at that moment).
  \item Ragno-Gioia:\textsuperscript{\ref{fn:Pisa2016}} the smartphone's lens was partially obscured, and the video had to be enhanced afterwards; there are huge difficulties in locating the grid, but the analysis goes well, except for moves 158 and 160, inverted.
  \item Telesca-Metta:\textsuperscript{\ref{fn:Pisa2016}} no problems at all, except for a grid recalibration at move 190 (when the tripod was bumped).
  \item Martinelli-van den Busken:\footnote{~\url{http://www.europeangodatabase.eu/EGD/Tournament_Card.php?&key=T161112C}} no problems at all. By the way, at the end of the game Martinelli, very disappointed for his last moves, resigned and hit the goban in frustration; in doing he so moved around several stones, prompting the program to detect two more moves, of course not to be recorded.
\end{enumerate}

On the whole, the analyses went rather well, especially because the number of frames under scrutiny was huge (more than 200,000): there were almost no errors detecting the stones while the grid, given the resolution was high enough, could be correctly located even when half of the goban's surface was covered by stones.

Of course some problems need to be solved: when the pace is too fast some stones may not be immediately detected at 480p, and even at 720p some difficulties may remain. Also, the grid is still difficult to locate at 480p. Furthermore, the program has not been tested under extreme situations, such as players moving stones around (on purpose or not), forgetting to remove captured ones, or even covering part of the goban. Although uncommon, such situations could nonetheless arise, and the program will have to deal with them and get on with the analysis, given the game has not been stopped. These are the issues we will try to improve in the next future:
\begin{itemize}
  \item the algorithm employed in detecting the stones, in order to locate those ones played too out of centre;
  \item the algorithm employed in detecting the grid, in order to let it work at low resolutions or with the goban almost covered by stones;
  \item also, we'll design a new algorithm, slow but more accurate than the one we employ in real-time, capable of checking and possibly correcting any wrong position that should appear, no matter the reason. This algorithm will work in a separate thread, without disturbing the main program's flow.
\end{itemize}

As shown, albeit still being under development, our program is already obtaining encouraging results in real life tests: handling those last three issues (the last one above all) will likely let us reach the ultimate goal of a truly unattended reconstruction of the whole move sequence of a Go game.

\bibliography{References}
\bibliographystyle{alpha}
\addcontentsline{toc}{section}{References}

\end{document}